\documentclass[conference]{IEEEtran}
\IEEEoverridecommandlockouts

\usepackage{cite}
\usepackage{amsmath,amssymb,amsfonts}
\usepackage{algorithmic}
\usepackage{graphicx}
\usepackage{subcaption}
\usepackage{textcomp}
\usepackage{csquotes}
\usepackage{xcolor}
\usepackage{booktabs}
\def\BibTeX{{\rm B\kern-.05em{\sc i\kern-.025em b}\kern-.08em
    T\kern-.1667em\lower.7ex\hbox{E}\kern-.125emX}}
\begin{document}

\title{Semantically Enriching Investor Micro-blogs for Opinion-Aware Emotion Analysis: A Practical Approach\\
\thanks{This work was conducted with the financial support of Research Ireland under Grant Number SFI/12/RC/2289\_P2 (Insight\_2)}
}

\author{\IEEEauthorblockN{Gaurav Negi}
\IEEEauthorblockA{\textit{Data Science Institute} \\
\textit{University Of Galway}\\
Galway, Ireland \\
}
\and
\IEEEauthorblockN{Paul Buitelaar}
\IEEEauthorblockA{\textit{Data Science Institute} \\
\textit{University Of Galway}\\
Galway, Ireland \\
}
}

\maketitle

\begin{abstract}
While sentiment analysis is the staple of financial NLP, capturing the nuances of ‘why’ behind that sentiment remains a challenge. There have been attempts to address this by analysing investor emotions alongside sentiment; however, this does not provide the additional granularity required to understand the target of the emotion/sentiment.  We address this by augmenting the StockEmotions dataset with semantically structured opinion graphs, which provide granular semantic depth to the existing sentiment and emotion labels. Using a declarative LLM pipeline, we augment the StockEmotions dataset with opinion graphs for each sentence, derived from 10,000 comments collected from StockTwits. In addition, we study the effect of introducing opinion semantics on baseline classifiers using Graph Neural Networks (GNNs).  Our analysis demonstrates that incorporating opinion semantics improves classification performance across different emotional spectrums. 
\end{abstract}


\section{Introduction}
The study of investor sentiment as a behavioural signal has always been a staple of financial decision-making. With the rise of micro-blogs and social media platforms, the target for these signals has shifted from surveys to these digital platforms, owing to reduced latency and increased scale \cite{micro_blog_1,micro_blog_2,behavioufin_2019}.

Despite sustained interest in investor sentiment, its direct impact on prices is often neutralised by arbitrageurs, as per the efficient market hypothesis \cite{sentiment_noeffect_2014, emh_1970}. However, sentiment exerts strong cross-sectional effects across various investment vehicles \cite{crosssection_2006}, suggesting that simple polarity is an insufficient metric. Consequently, recent studies have begun to analyse the nuanced intricacies of investor emotion \cite{emo_limited_2018, lee2023stockemotions} by assigning utterances to richer categories than binary sentiment. Yet, a gap remains: the existing representational layer in the literature has not yet transitioned from these emotional categories to fine-grained opinions, which would integrate specific targets with deep semantic and contextual information.

In this work, we address this representational research gap and augment the StockEmotions \cite{lee2023stockemotions} dataset with fine-grained opinion representation based on the Unified Opinion Concepts (UOC) ontology \cite{uoc-2025}. The UOC ontology formalises and unifies diverse opinions and fine-grained sentiment representations. The process of augmentation is carried out with a Large Language Model (LLM)- driven pipeline, which is capable of reasoning and few-shot learning, i.e., not requiring parametric learning \cite{few_shot_learners_2020, llm_reasoner_2022}.

The augmented dataset merges human-annotated emotional categories with fine-grained opinion semantics derived from UOC. Grounded in the affective framework \cite{munezero2014they}, we investigate the influence of opinion semantics on emotion prediction. In our analysis, impulse-driven emotions serve as the dependent variables, while information-driven opinions and the original text act as predictors. Using this dataset, we address the following research questions:
\begin{enumerate}
    \item [RQ1.] Can the semantically rich representation of opinion assist in improving the accuracy of predicting investor's emotional states?
    \item [RQ2.] How to effectively integrate opinion signals in emotional classifier preserving the semantic information?
\end{enumerate}

The premises of our research questions also serve to verify the usefulness of the augmentation, where improvements in prediction performance are assessed through statistically significant differences in model outputs. Our contributions can be summarised as:
\begin{itemize}
    \item We augment the StockEmotion dataset with detailed opinions based on the UOC ontology using an LLM-driven annotation pipeline.  
    \item We propose integrating UOC opinion semantics into a pre-trained language model classifier using Graph Neural Networks. 
    \item We evaluate state-of-the-art LLMs for emotion classification on financial texts.
\end{itemize}

\section{Related Work}
The prediction of sentiment has become a staple of the finance domain, where NLP has been widely used for predictive modelling of investor sentiment \cite{mirco_blog_senti_2016, fin_senti_2017} and investor emotions \cite{lee2023stockemotions, Duxbury21092020}. These sentiments in early research lacked contextual information, as is customary in NLP opinion mining tasks, both in theorised \cite{liu2010sentiment, Liu2017} and operationalised \cite{pontiki-etal-2016, peng_2020} forms. This shortcoming was addressed by fine-grained sentiment shared tasks for the finance domain \cite{cortis-etal-2017-semeval}; however, their formulation lacks the semantic details provided by ontological specification of Unified Opinion Concepts \cite{uoc-2025}.

Emotional states have been expressed in categorical classes following Ekman's \cite{ekman_1992} six categories and Plutchik's eight categories in various theoretical and analytical fields. More recently, in NLP, emotional classification has been operationalised in a more refined set of 27 categories \cite{goemo_2020}. There have been numerous emotion-prediction studies in NLP \cite{kusal2023systematic} and affective computing \cite{graterol2021emotion}; however, to the best of our knowledge, the interaction between opinion semantics and emotions remains underexplored. This is the research gap we address by investigating the interaction between the rich semantics of opinions, as specified by the UOC, and investor emotions.

A key prerequisite for our experiments is the availability of both emotion and detailed opinion annotations along with semantic information. We choose a dataset with human annotations for emotion, as this is our observed variable, and then use an LLM-based pipeline to augment the data with semantically rich UOC opinion labels. LLMs have been investigated for annotating various subjective tasks, including span annotations \cite{annot_spans_2025}, argument quality annotations \cite{argue_annot_2024}, and propaganda span annotations \cite{annotate_propaganda_2024}, with findings arguing in favour of using LLMs as independent or additional annotators.

\section{Dataset}

\begin{table}[htbp]
    \centering
    \begin{tabular}{ l r r r }
        \toprule
        \textbf{Emotion} & \textbf{Train (\%)} & \textbf{Dev (\%)} & \textbf{Test (\%)} \\
        \midrule
        Optimism   & 16.24 & 16.20 & 16.30 \\
        Anxiety    & 13.74 & 13.30 & 13.40 \\
        Excitement & 13.65 & 14.80 & 14.60 \\
        Disgust    & 12.96 & 12.10 & 12.10 \\
        Belief     & 9.10  & 9.10  & 8.90  \\
        Ambiguous  & 8.72  & 8.60  & 8.70  \\
        Amusement  & 8.15  & 8.30  & 8.30  \\
        Confusion  & 6.11  & 6.00  & 6.00  \\
        Anger      & 3.86  & 3.90  & 3.80  \\
        Panic      & 3.00  & 3.30  & 3.10  \\
        Surprise   & 2.39  & 2.40  & 2.90  \\
        Depression & 2.08  & 2.00  & 1.90  \\
        \midrule
        \textbf{Total Count} & \textbf{8000} & \textbf{1000} & \textbf{1000} \\
        \bottomrule
    \end{tabular}
    \caption{Dataset Breakdown of Emotions}
    \label{tab:emotion_splits}
\end{table}
The primary objective of our work is to analyse the interaction between investors' emotional states and the semantics of opinions, we conduct our experiments on the StockEmotions dataset, which provides human-annotated target labels for emotional states. The data is collected from the microblogging website Stocktwits. It is an investor-centric social media platform where the primary topic of interest is always centred on finance. The dataset breakdown is discussed in table \ref{tab:emotion_splits}.


\section{Methodology}
\begin{figure*}[htbp]
    \centering
    \begin{subfigure}[b]{.38\textwidth}
        \centering
        \includegraphics[width=\textwidth]{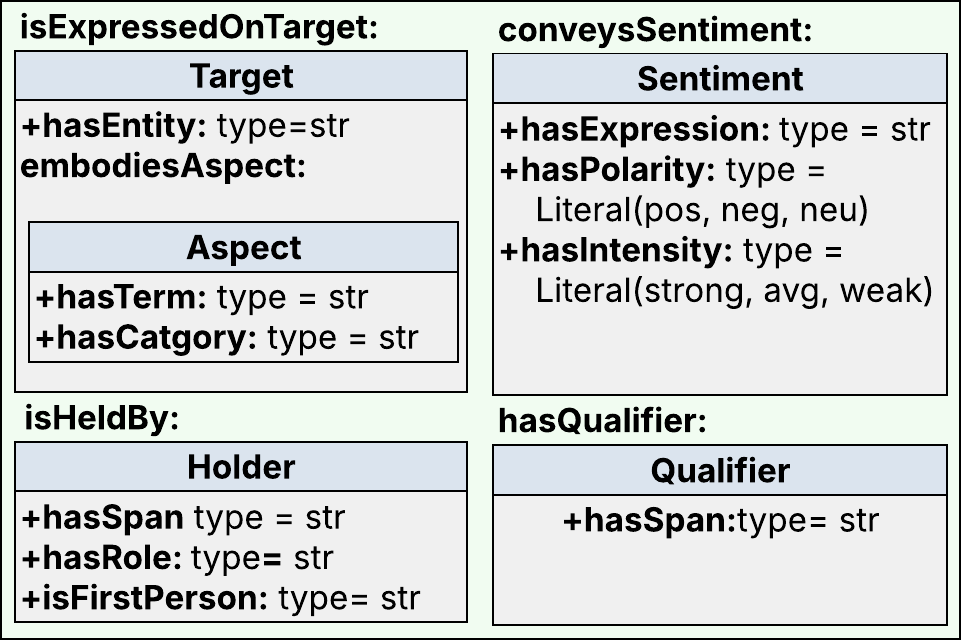}
        \caption{Opinion Schema based of UOC}
        \label{fig:uoc_schema}
    \end{subfigure}
    \hfill 
    \begin{subfigure}[b]{.55\textwidth}
        \centering
        \includegraphics[width=\textwidth]{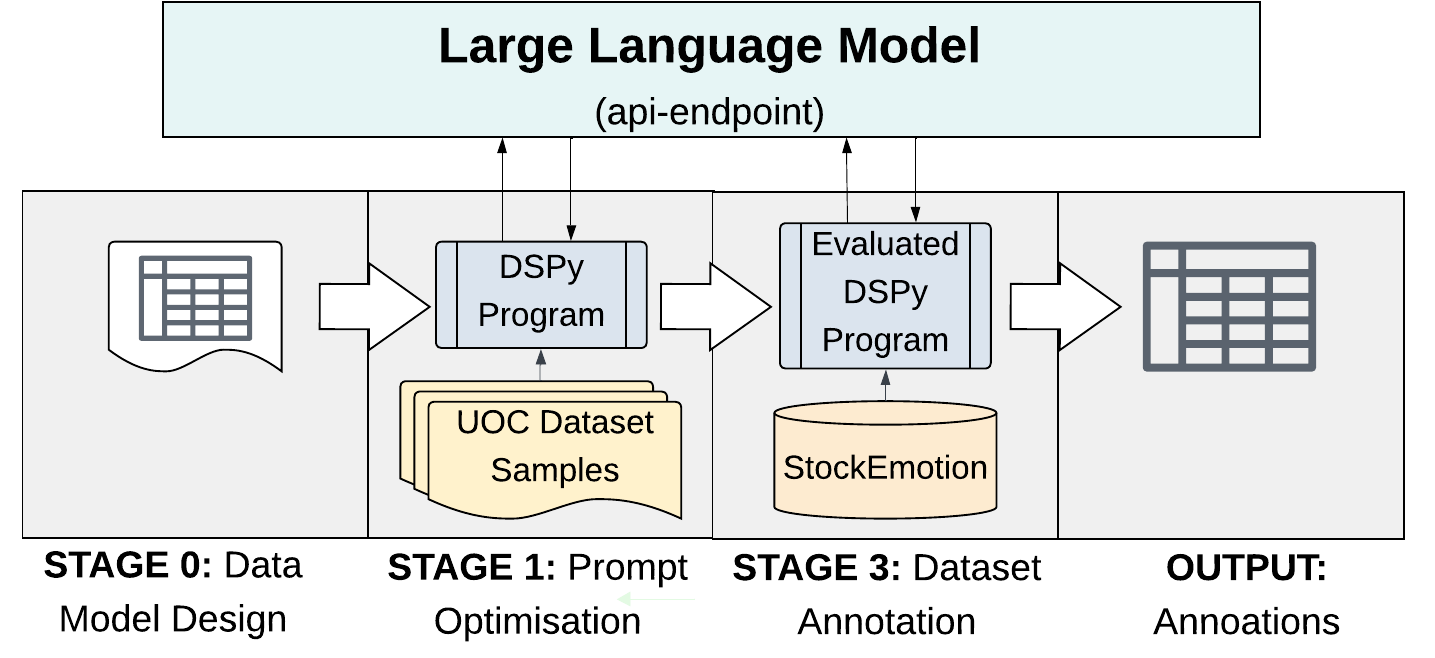}
        \caption{Declarative Augmentation Pipeline}
        \label{fig:annot_pipeline}
    \end{subfigure}
    
    \label{fig:augmentation_pipe}
    \vspace{-10pt}
\end{figure*}

\subsection{Generating Opinion Annotations with LLM} 

\noindent \textbf{Opinion Representation}: We augment our dataset with fine-grained opinion formulation introduced by \cite{uoc-2025} and illustrated in Fig \ref{fig:uoc_schema}, which is a comprehensive and semantically rich description of the opinion. The augmentation introduces several new concepts which describe not just the sentiment but also the contextual elements present. These concepts are described below with an example used to aid the description:

\begin{displayquote}
    \textbf{Example} \textit{I had hoped for better battery life, as it had only about 2-1/2 hours doing heavy computations (8 threads using 100\% of the CPU)}
\end{displayquote}
\begin{itemize}
    \item \textbf{Sentiment}: Encapsulates the underlying feelings expressed in an opinion. It encapsulates descriptive properties, namely sentiment polarity (positive, negative, or neutral), sentiment intensity (strong, average, or weak), and sentiment expression in the text. In the example above: sentiment is specified by (i) Sentiment Expression: \textit{``hoped for better''}, (ii) Sentiment Polarity: \textit{negative} and (iii) Sentiment Intensity: \textit{average}
    \item \textbf{Target}: The subjective information on which an opinion is expressed. It has aspect elements and a coarse-grained Target Entity. 
    In the example, the opinion is expressed regarding the Target entity: \textit{Battery}.
    
    \item \textbf{Aspect}: Aspect describes the part and attribute of the Target Entity on which the sentiment is expressed. It is composed of an explicitly instantiated aspect term in the text and a more coarse-grained property, a category, that expresses the attribute of the entity discussed. The Aspect of opinion in the Example can be expressed into the Aspect Term: \textit{``battery life''} and Aspect Category: \textit{Operation\_Performance}
    
    \item \textbf{Holder}: A holder is the individual or organisation expressing an opinion. In the example, the holder is identified by the span \textit{``I''}.
    
    \item \textbf{Qualifier}: Specifies the group/subgroup to which the opinion to which the opinion pertains. In the above example, the affected people are from the group that performs heavy computations. Therefore \textit{``doing heavy computations''} is the Qualifier.
\end{itemize}

\noindent \textbf{LLM-based Augmentation Pipeline}: The opinions to augment our dataset are generated following the pipeline introduced by \cite{kbc_dspy_2025} that follows the paradigm of ``programming LLM'' \cite{dspy_2024} rather than using naive prompts, which are known to be sensitive \cite{prompt_sensitive_2024}. Synthesising the prompt also minimises spurious interactions between the prompt and model selection as confounding variables, since the prompt is now automatically generated. The annotation pipeline utilises the Multi-prompt Instruction Proposal Optimiser (MIPRO) \cite{mipro_v2_2024} to jointly optimise both the instructions and demonstrations, i.e. the In-Context Examples. It operates by generating prompt proposals P, then performing a Bayesian-inspired lookup over the combination of P and ICL examples using an evaluation metric. The annotation pipeline is illustrated in Fig. \ref{fig:annot_pipeline}.

\subsection{Baseline Model Architecture}
\noindent The input sequence is text only $T$, which is transformed to the embedding space using encoder only Transformer models $H \in \mathbb{R}^{|T| \times d}$. The final representations are obtained as features corresponding to the $[CLS]$ token for Bert model and model specific pooling for others which essentially reduces the dimension to $H_{seq} \in \mathbb{R}^{d}$, for sequence level prediction with the classification head $z=WH_{seq}+b$ where $z\in\mathbb{R}^{C}$ and C denotes the number of classes.

\subsection{Injecting Semantics of Opinion Representation}

\begin{figure}[htbp]
        \centering
        \includegraphics[width=0.80\linewidth]{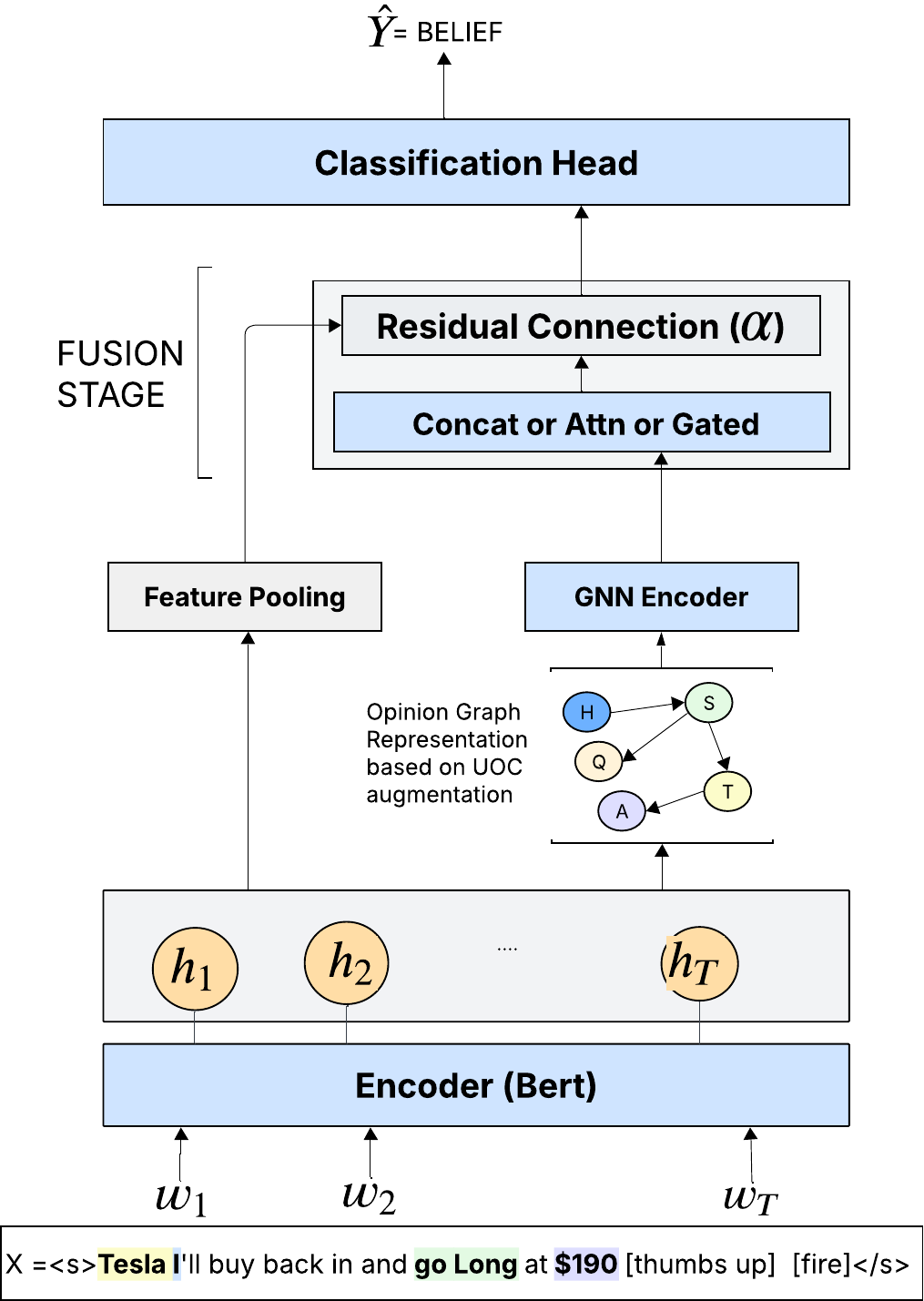}
        \caption{Model Architecture}
        \label{fig:gnn_aug}
        \vspace{-35pt}
\end{figure}

\noindent We treat the task as a multi-class classification task. We conduct the experiments with transformer-based encoder only architectures. We use Graph Neural Networks (GNN) to encode the semantics of the opinions as features which are then fused with the token level representations of the baseline architecture, as illustrated in fig 
\ref{fig:gnn_aug}. \medskip

\noindent \textbf{STAGE-1: Opinion Sub-graph Extraction} \smallskip

\begin{figure}[htbp]
        \centering
        \includegraphics[width=.8\linewidth]{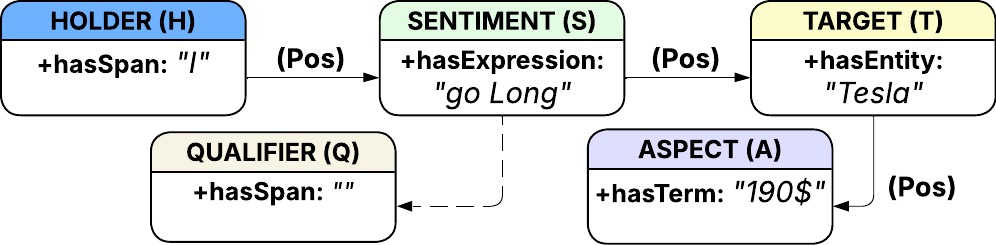}
        \caption{Sub-graph for : \textit{``Tesla I'll buy back in and go Long at \$190 [thumbs up]  [fire]''}}
        \label{fig:graph_eg1}
\end{figure}

\noindent The opinion graphs extracted for the input text during the data augmentation step comprise: (i) the spans in text, (ii) inferred characteristics associated with them, and (iii) Relations between them. In this work, our model is based entirely on the text sequence, so we construct graphs using only spans from the text and their relations. The only additional information utilised by the model is the semantics and sentiment polarity provided during the augmentation step. This schema for message passing within the sub-graph is illustrated in Figure \ref{fig:graph_eg1}.

The illustrated graph is expressed as: $G = (V, E, \mathcal{E})$, where $V = \{v_{h}, v_{S}, v_{t}, v_{q}, v_{a}\}$ represent the set of nodes corresponding to the opinion holder, source target, qualifier and aspect, respectively as illustrated in Figure \ref{fig:graph_eg1}, $E$ is the sequence of edge-indices that encodes the flow of messages and $\mathcal{E}$ is the edge attribute that encodes the sentiment polarity.

The node features are initialised from transformer hidden states $H \in \mathbb{R}^{|T| \times d}$. For a span $r$ (e.g. the holder), node representation is extracted as $ h_i = \mu(H[r])$, where $\mu(.)$ denotes the mean-pooling operator. The neighbourhood of node $i$ is defined as $\mathcal{N}(i)=\{j|(i,j)\in E\}$

The number of graphs corresponding to each sentence is not fixed. We suppose there are $m$ opinion graphs $\{G_1, G_2, …, G_m\}$ per text example, each graph being $G_m = (V_m, E_m, \mathcal{E}_m)$.\medskip

\noindent \textbf{STAGE-2: Message Passing over Opinion Sub-graphs} \smallskip

\noindent The sub-graphs created in the section serve as input to the Graph Attention Layer (GATv2) \cite{gat_v2_2022}, which learns semantic representations in a parametric form by aggregating information from the neighbourhood nodes of $i$. For node $i \in V$

$$h'_i = \sum_{j\in N(i)\cup\{i\}}\alpha_{ij} \Theta_t h_j$$ With attention coefficients $\alpha$:
$$\alpha_{i,j}=\frac{\text{exp}(a^{T}\text{LeakyReLU}(\Theta_s h_i + \Theta_th_j))}{\sum_{k\in\mathcal{N}(i) \cup\{i\}} \text{exp}(a^T \text{LeakyReLU }(\Theta_s h_i + \Theta_th_k))}$$
Where $\alpha_{i,j}$ follows the specification that incorporates edge attributes $e_{ij}$ to encode sentiment polarity of the opinion as a feature. We use a multi-head GAT with K attention heads to independently aggregate neighbourhood information, $H'_m = \{h_1', h_2', ..., h'_{|V_m|}\}$, where each $h_i'$ is the concatenation of $K$ heads: $ h'_i=\bigoplus_{k=1}^{K} \text{GATv2}^{(k)}(G_m)$.
\medskip

\noindent \textbf{STAGE-3: Mapping Graph Space to Sequence Space}\smallskip

\noindent The node-level features $h'_i$ for each graph  $G_m$ are sum-pooled to provide graph-level representations: $X'_m = \sum_{v_i\in Vm}h'_{i}$. To ensure these GNN features are compatible with sequence-level representation, we align the $M$  extracted graph features with their corresponding $S$ sentences using a mapping function $\mathcal{M}$, which assigns each opinion sub-graph to its parent sentence, i.e. $\mathcal{M}: \{1, 2, ..., M\} \rightarrow \{1, 2, ...S\}$. The final opinion-informed semantically enriched features for a given sentence $s$ are calculated as the mean of its associated graph projections: 
$$H_s^{\mathcal{G}} = \text{MEAN}(\{X'_m|\mathcal{M}(m)=s\})$$ \medskip

\noindent \textbf{STAGE-4: Fusion of Features}\smallskip

\noindent Prior to the classification head, the baseline features $H_{seq}$ and semantic features $H_s^{\mathcal{G}}$ are fused together via fusion function $H_f = \text{Fuse}(H_{seq},H^\mathcal{G}_{s})$. We experiment with three types of feature fusion by searching over them as a hyper-parametric variation:

\begin{enumerate}
    \item Concatenation (cat): These features are concatenated into a single vector $H_{seq}||H_s^\mathcal{G} \in \mathbb{R}^{2d}$ and projected back to $d$ using a linear layer.
   $$H_f = W_g[H_{seq}||H_s^\mathcal{G}]+b_g$$
   A non-selective combination of both features characterises this type of fusion.
   \item Gated Fusion (gate): To allow the model to weight the importance of text versus opinion semantics dynamically, we implement a gating mechanism: $g=\sigma(\Theta_g[H_{seq}||H_s^\mathcal{G}]+b_g)$. The final representation is the weighted combination: $$H_f = g \odot H_{seq} + (1-g)\odot H_s^{\mathcal{G}}$$
   \item Attention Fusion (attn): To model the interactivity between the two feature spaces, we use a dot-product attention mechanism where $Q=H_s^{\mathcal{G}}$ and $K=V=H_{seq}$: $$H_f = \text{softmax}\bigg(\frac{Q \cdot K^{T} }{\sqrt{d}}\bigg)\cdot V$$
\end{enumerate}
Finally, the fused representation goes through a residual stage, where the original sequence features $H_{seq}$ are added back to preserve the textual context. This residual stage is regulated by a scaling hyper-parameter $\alpha_{res}$. $$H_R = H_{seq}+\alpha_{res} \cdot H_{f}$$

\begin{table*}[htbp]
\centering

\resizebox{\textwidth}{!}{%
\begin{tabular}{lcccccccccccc|c}
\toprule
\textbf{Model} & \textbf{depre.} & \textbf{panic} & \textbf{surpr.} & \textbf{anger} & \textbf{confu.} & \textbf{amuse.} & \textbf{ambig.} & \textbf{belief} & \textbf{disgu.} & \textbf{anxiety} & \textbf{excit.} & \textbf{optim.} & $\text{F}_{1(macro)}$ \\
\midrule
RoBERTa           & 40.00 & \underline{28.57} & 12.50 & \textbf{\underline{46.15}} & \textbf{\underline{61.29}} & 38.50 & 12.72 & \textbf{\underline{33.71}} & 39.43 & \underline{40.42} & \underline{47.83} & \textbf{\underline{49.70}} & 37.57 \\
RoBERTa-GNN & \textbf{\underline{41.86}} & 25.39 & \underline{26.92} & 42.85 & 58.82 & \textbf{\underline{41.05}} & \underline{14.41} & 23.89 & \textbf{\underline{45.36}} & 40.14 & 46.05 & 49.26 & \textbf{\underline{38.01}} \\\hline
BERT              & \underline{27.69} & \textbf{\underline{35.82}} & \textbf{\underline{33.96}} & 34.48 & 54.71 & 22.72 & 14.28 & 25.31 & 35.43 & 36.46 & 42.58 & 35.63 & 33.25 \\
BERT-GNN      & 18.18 & 34.04 & 30.43 & \underline{35.48} & \underline{59.45} & \underline{40.64} & \textbf{\underline{23.12}} & \underline{30.02} & \underline{40.86} & \textbf{\underline{50.19}} & \textbf{\underline{48.25}} & \underline{43.67} & \underline{37.88} \\\hline
GPT-5-MINI        & \underline{30.76} & 21.05 & \underline{23.8} & 35.08 & 32.18 & 19.74 & 2.00 & 4.00 & 3.149 & 8.75 & 37.28 & 30.21 & \underline{20.67} \\
GPT-5             & 24.00 & 15.38 & 12.50 & \underline{35.16} & \underline{36.36} & \underline{21.34} & 3.50 & \underline{6.28} & \underline{6.41} & 13.33 & \underline{38.44} & \underline{33.54} & 20.52 \\
Qwen-3.5-35B      & 22.85 & \underline{21.27} & 9.70 & 33.34 & 35.84 & 19.87 & \underline{6.20} & 6.15 & 2.98 & \underline{22.34} & 32.74 & 32.44 & 20.48 \\
\bottomrule
\end{tabular}%
}
\caption{$F_1$ Scores denoting model performance Across emotion Categories. The GNN fusion was finalised with hyper-parameter search, specifically Gated fusion for Roberta and Concatenation fusion for Bert.}
\label{tab:emotion_results}
\vspace{-15pt}
\end{table*}

\begin{figure*}[h]
        \centering
        \includegraphics[width=\linewidth]{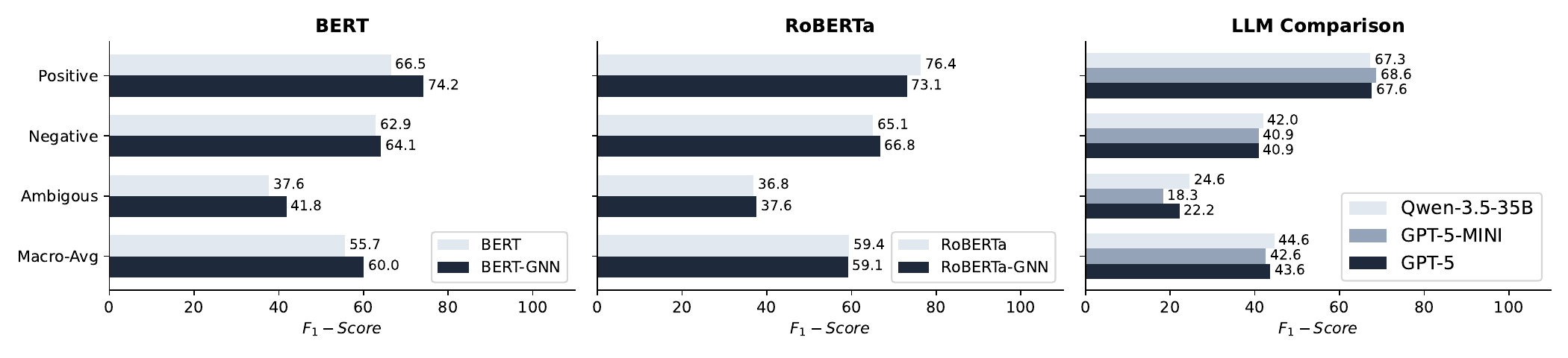}
        \caption{Aggregating performance on valence with the mapping provided with Go-Emotions \cite{goemo_2020}.}
        \label{fig:valence}
        \vspace{-10pt}
\end{figure*}

\section{Experimental Setup}
SFT experiments (BERT \cite{bert-devlin-2019}, RoBERTa \cite{roberta-2019}) used an NVIDIA RTX 4090 (24GB), while LLM data augmentation (Qwen3.5-35B \cite{qwen_35}) utilised an NVIDIA A40 (48GB). We hosted an 8-bit quantised Qwen3.5-35B via VLLM and implemented semantic message-passing using GATv2 (\textit{pytorch\_geometric}). Using Bayesian Sweeps, searched through the model hyper-parameters: batch sizes $\in \{8, 16, 32, 64\}$, GATv2 output dimensions $\in \{384, 256, 192, 96\}$, attention heads $\in \{2, 3, 4, 6, 8\}$, fusion types $\in \{\text{'cat', 'attn', 'gated'}\}$, and residual factor $\alpha_{res} \in \{0.25, 0.50, 0.75, 1\}$.

\section{Results}
The performance of the evaluated models across the twelve emotion categories is summarised in Table \ref{tab:emotion_results}. To isolate the impact of the proposed graph-based architectures, we compared our models (Roberta-GNN and Bert-GNN) directly against our re-implemented baselines of the standard Roberta and Bert models \cite{lee2023stockemotions}. We also benchmarked these against several large generative language models (GPT-5-MINI, GPT-5, and Qwen-3.5-35B) evaluated in a zero-shot classification setting. Overall, the fine-tuned encoder models outperformed the larger generative models on this specific classification task in the finance domain.

\subsection{Statistical Significance of Semantic Interventions}
The integration of Graph Neural Networks (GNNs) significantly improved the BERT model, with our BERT-GNN variant achieving an $F_{1\text{-macro}}$ score of 37.88, up from 33.25 and yielding an improvement of over 4.5 points ($p \approx 0.001$). It excelled in categories like \textit{anxiety} (50.19), \textit{excitement} (48.25), and \textit{ambiguity} (23.12).

For  RoBERTa, the GNN addition led to a slight increase in performance, with the score rising from 37.57 to 38.01. Although not statistically significant via McNemar’s test, the Stuart-Maxwell test indicated a significant change in label assignments ($p \approx 3.68 \times 10^{-7}$), achieving the highest scores in \textit{depression} (41.86) and \textit{disgust} (45.36).

\subsection{Category-Specific Observations}
Certain emotions proved notably easier to classify, ``Confusion'' and ``Optimism'' consistently yielded the highest scores among the top-performing encoders, peaking at 61.29 and 49.70, respectively. However, "Ambiguity" and "Belief" were universally challenging, even the GNN-enhanced models struggled to surpass F1 scores in the low $20$s and $30$s for these categories. Notably, the large generative models exhibited a near-total failure in detecting "Ambiguity," "Belief," and "Disgust" in their zero-shot configurations, with performance dropping into the single digits. 

\subsection{Analysis}
To evaluate model robustness, we mapped predictions to Ekman’s six basic emotions in Table \ref{tab:aggregated_results}  and a three-class valence grouping (Positive, Negative, Ambiguous) in Figure \ref{fig:valence}, along with their macro-averaged F1 score. The GNN architecture significantly enhanced the Bert model's performance, with BERT-GNN achieving the highest valence Macro-F1 (60.04), primarily due to gains in Positive (+7.69) and Ambiguous (+4.18) classifications. In RoBERTa where we observed a redistribution of labels, we see that the score decresed for the Positive emotions but there was improvement in the prediction of both negative and ambigours emotions. 
In contrast, generative LLMs showed strong zero-shot performance in identifying Positive emotions (F1 scores near 68) but struggled with Negative and Ambiguous classifications, impacting their overall performance.

\begin{table}[htbp]
\centering
\renewcommand{\arraystretch}{1.2} 
\setlength{\tabcolsep}{2pt}      
\resizebox{0.95 \columnwidth}{!}{%
\begin{tabular}{l *{6}{c}c} 
\toprule
\textbf{Model} & \textbf{Anger} & \textbf{Disgu.} & \textbf{Fear} & \textbf{Joy} & \textbf{Sadne.} & \textbf{Surpr.} & \boldmath{$\text{F}_{1(mac)}$} \\
\midrule
RoBERTa     & \textbf{46.15} & 39.43 & 49.69 & \textbf{76.36} & 40.00 & 36.84 & 48.05 \\
RoBERTa-GNN & 42.85 & \textbf{45.36} & 48.66 & 73.08 & \textbf{41.86} & 37.58 & \textbf{48.23} \\\midrule
BERT        & 34.48 & 35.43 & 43.18 & 66.51 & 27.69 & 37.63 & 40.82 \\
BERT-GNN    & 35.48 & 40.86 & \textbf{52.98} & 74.20 & 18.18 & \textbf{41.81} & 43.92 \\
\midrule
GPT-5-MINI  & 35.08 & 3.14  & 27.64 & 68.64 & 30.76 & 18.34 & 30.60 \\
GPT-5       & 35.16 & 6.41  & 24.88 & 67.63 & 24.00 & 22.22 & 30.05 \\
Qwen-3.5-35B& 33.34 & 2.98  & 30.97 & 67.26 & 22.85 & 24.63 & 30.34 \\
\bottomrule
\end{tabular}%
}
\caption{Aggregated Performance Across Emotional Taxonomies.}
\label{tab:aggregated_results}
\vspace{-15pt}
\end{table}

\section{Conclusion}
This paper evaluates the effects of integrating the UOC ontology with a dataset using an LLM-driven pipeline for fine-grained emotion classification. By incorporating augmented semantics of opinions into graph-enhanced model architectures with an encoder, we compared results against standardised baselines and advanced LLMs. The findings reveal that semantically enhanced opinion-aware models, particularly those integrating GNNs into the BERT architecture, significantly improve accuracy. For RoBERTa, a graph-gating mechanism improves performance on complex emotions such as depression and disgust. Our analysis revealed limitations of large language models (e.g., GPT-5, Qwen-3.5-35B) in accurately identifying ambiguous and negative emotions, with task-specific encoders outperforming them, even in broader emotion taxonomies.

\bibliographystyle{IEEEtran}
\bibliography{references}
\end{document}